\documentclass[10pt]{article}
\usepackage{amsmath}
\usepackage{amsfonts}
\usepackage{latexsym}
\usepackage{graphicx}
\usepackage{url}
\usepackage{colortbl}
\usepackage{rotating}
\usepackage[font=normal,format=plain,labelfont={sf,bf},up,textfont=it,up]{caption}
\DeclareFontFamily{OT1}{pzc}{}
\DeclareFontShape{OT1}{pzc}{m}{it} {<-> s * [1.2] pzcmi7t}{}
\DeclareMathAlphabet{\mathpzc}{OT1}{pzc}{m}{it}
\newcommand{\db}{\mathpzc{d}_\mathcal{B}}
\newcommand{\R}{\mathbb{R}}

\begin{document}

\title{Fast, Linear Time Hierarchical Clustering using the Baire Metric}
\author{Pedro Contreras (1) and Fionn Murtagh (1,2) \\
(1) Department of Computer Science \\
Royal Holloway, University of London \\
Egham TW20 0EX, UK \\
(2) Science Foundation Ireland \\
Wilton Park House, Wilton Place, Dublin 2, Ireland \\
Email: pedro@cs.rhul.ac.uk, fmurtagh@acm.org}
\date{}
\maketitle

\begin{abstract}
The Baire metric induces an ultrametric on a dataset and is 
of linear computational complexity, contrasted with the standard
quadratic time agglomerative hierarchical clustering algorithm.  
In this work we evaluate empirically 
this new approach to hierarchical clustering. 
We compare hierarchical clustering based on the Baire metric with 
(i) agglomerative hierarchical clustering, in terms of algorithm properties; 
(ii) generalized ultrametrics, in terms of definition; and 
(iii) fast clustering through k-means partititioning, in terms of 
quality of results.  For the latter, we carry out an in depth 
astronomical study.  
We apply the Baire distance to spectrometric and photometric redshifts 
from the  Sloan Digital Sky Survey using, in this work, about half a 
million astronomical objects.   We want to know how well the 
(more costly to determine) spectrometric redshifts can predict the 
(more easily obtained) photometric redshifts, i.e.\ we seek to 
regress the spectrometric on the photometric redshifts, and we use
clusterwise regression for this.   
\end{abstract}


\section{Introduction}
\label{intro}

Our work has quite a range of vantage points, including the 
following.  Firstly, there is a particular distance between
observables, which happens to be also a ``strong'' or 
ultrametric distance.  Section \ref{section:baire} 
defines this.   This same section notes how the encoding
of data is quite closely associated with the determining of 
the distance.   

Next, in section \ref{genum} we take the vantage point of 
clusters, and of sets of clusters.   

Finally, in section \ref{subsection:alg} we wrap up on 
the hierarchy that is linked to the distance used, and to the 
set of clusters.  

So we have the following aspects and vantage points: distance, 
ultrametric, data encoding, cluster or set (and membership), 
sets of clusters (and their interrelationships), and hierarchical
clustering.   Those aspects and vantage points are discussed
in the first part of this article.  They are followed by case 
studies and applications in subsequent sections.  We have not, 
in fact, exhausted the properties and aspects of our new 
approach.  For example, among issues that we will leave for 
further in depth exploration are: p-adic number representation
spaces; and hashing, data retrieval and information obfuscation.

The following presents a general scene-setting where we introduce
metric and ultrametric,  we describe
some relevant discrete mathematical structures,
and we note some computational properties.  

\subsection{Agglomerative Hierarchical Clustering Algorithms}

A metric space $(X,d)$ consists of a set~$X$ on which is defined a 
distance function $d$ which assigns to each pair of points of 
$X$ a distance between them, and satisfies the following four axioms 
for any triplet of points $x, y , z$:

\begin{equation*} 
\mbox{A1: } \forall x, y \in X,  d(x,y) \geq 0 \mbox{ (positiveness)}
\end{equation*}

\begin{equation*}
\mbox{A2: } \forall x, y \in X, d(x,y) = 0 \mbox{ iff } x = y \mbox{ (reflexivity)}
\end{equation*}

\begin{equation*}
\mbox{A3: } \forall x, y \in X, d(x,y) = d(y,x)  \mbox{ (symmetry)}
\end{equation*}

\begin{equation*}
\mbox{A4: } \forall x, y, z \in X, d(x,z) \leq d(x,y) + d(y,z)  \mbox{ (triangle inequality)}
\end{equation*}

When considering an ultrametric space we need to consider 
the strong triangular inequality or ultrametric inequality defined as: 

\begin{equation*}
\mbox{A5: } d(x,z) \leq max~\{d(x,y), \ d(y,z)\} \mbox{ (ultrametric inequality)} 
\end{equation*}
and this in addition to the positivity, reflexivity and symmetry properties 
(properties A1, A2, A3) 
for any triple of point  $x, y, z \in X$.  

If $X$ is endowed with a metric, then this metric can be 
mapped onto an ultrametric.  In practice, endowing  $X$ 
with a metric can be relaxed to a dissimilarity.  An often 
used mapping from metric to ultrametric 
is by means of an agglomerative hierarchical clustering 
algorithm.  A succession of $n - 1$ pairwise merge steps takes place by making 
use of the closest pair of singletons and/or clusters at each step.  Here 
$n$ is the number of observations, i.e.\ the cardinality of set $X$.  
Closeness between
singletons is furnished by whatever distance or dissimilarity is in use.
For closeness between singleton or non-singleton clusters, we need to 
define an inter-cluster distance or dissimilarity.  This can be defined
with reference to the cluster compactness or other property that we wish 
to optimize at each step of the algorithm.   In terms of advising a 
user or client, such a cluster criterion, motivating the inter-cluster
dissimilarity, is best motivated in turn by the data analysis application
or domain.  

Since agglomerative hierarchical clustering requires consideration of 
pairwise dissimilarities at each stage it can be shown that even in the 
case of the most 
efficient algorithms, e.g.\ those based on reciprocal nearest neighbors
and nearest neighbor chains \cite{Murtagh85-1}, $O(n^2)$ or quadratic 
computational time is required.  The innovation in the work we present here
is that we carry out hierarchical clustering in a different way such that
$O(n)$ or linear computational time is needed.  As always in computational 
theory, these are worst case times.  

A hierarchy, $H$,
is defined as a binary, rooted, node-ranked tree, also
termed a dendrogram \cite{benz,john,Lerman81,Murtagh85-1}.
A hierarchy defines a set of embedded subsets of a given set of objects
$X$, indexed by the set $I$.
These subsets are totally ordered by an index function $\nu$, which is a
stronger condition than the partial order required by the subset relation.
A bijection exists between a hierarchy and an ultrametric space.

Let us show these equivalences between embedded subsets, hierarchy, and
binary tree, through the constructive approach of inducing $H$ on a set
$I$.

Hierarchical agglomeration on $n$ observation vectors with indices
$i \in I$ involves
a series of $1, 2, \dots , n-1$ pairwise agglomerations of
observations or clusters, with properties that follow.  

In order to 
simplify notation, let us use the index $i$ to represent also the 
observation, and also the observation vector.  Hence for $i = 3$ and 
the third -- in some sequence -- observation vector, $x_i = x_3$, we
will use $i$ to also represent $x_i$ in such a case.   

A hierarchy 
$H = \{ q | q \in 2^I \} $ such that (i) $I \in H$, (ii) $i \in H \ \forall 
i$, and (iii) for each $q \in H, q^\prime \in H: q \cap q^\prime \neq 
\emptyset \Longrightarrow q \subset  q^\prime \mbox{ or }  q^\prime 
 \subset q$.  Here we have denoted the power set of set $I$ by $2^I$.
An indexed hierarchy is the pair $(H, \nu)$ where the positive
function defined on $H$, i.e., $\nu : H \rightarrow \R^+$, satisfies:
$\nu(i) = 0$ if $i \in H$ is a singleton; and (ii)  $q \subset  q^\prime 
\Longrightarrow \nu(q) < \nu(q^\prime)$.  Here we have denoted the
positive reals, including 0, by $\R^+$.
Function $\nu$ is the agglomeration
level.  Take                
$q \subset q''$
 and $q^\prime \subset q''$, and let $q''$ be the lowest level cluster for
which this is true. Then if we define $D(q, q^\prime) = \nu(q'')$, $D$ is
an ultrametric.  

In practice, we start with a Euclidean or alternative
dissimilarity, use some criterion such as minimizing the change in variance
resulting from the agglomerations, and then define $\nu(q)$ as the
dissimilarity associated with the  agglomeration carried out.

\section{Baire or Longest Common Prefix Distance}
\label{section:baire}

Agglomerative hierarchical clustering algorithms are constructive 
hierarchy-constructing algorithms.  Such algorithms have the aim 
 of mapping data into an ultrametric space, or
searching for an ultrametric embedding, or ultrametrization~\cite{Rooij78}. 

Now, inherent ultrametricity leads to an identical result with most 
commonly used agglomerative criteria~\cite{Murtagh85-1}. Furthermore, data 
coding can help greatly finding how inherently ultrametric data 
is~\cite{Murtagh04}.  In certain respects the hierarchy determined by the 
Baire distance can be viewed as a particular coding of the data because
it seeks longest common prefixes in pairs of (possibly numerical) strings.  
We could claim that determining the longest common prefix is 
a form of data compression because we can partially 
express one string in terms of another.  

\subsection{Ultrametric Baire Space}
\label{subsection:baire}

A Baire space consists of countably infinite sequences with a metric defined in terms of the longest common prefix: the longer the common prefix, the closer a pair of sequences. What is of interest to us here is this longest common prefix metric, which we call the Baire distance~\cite{murtagh08, Contreras07}. 

Consider real-valued or floating point data (expressed as a string of digits
rather than some other form, e.g.\ using exponent notation).   
The longest common prefixes at issue are those of precision of any value.  
For example, let us consider two such values, $x_{i}$ and $y_{j}$, with
$i$ and $j$ ranging over numeric digits.   When the context easily allows 
it, we will call these $x$ and $y$. 

Without loss of generality we take $x$ and $y$ to be real-valued and 
bounded by 0 and 1. 

Thus we consider ordered sets $x_{k}$ and $y_{k}$ for $k \in K$. In line with our notation, we can write $x_{k}$ and $y_{k}$ for these numbers, with the set $K$ now ordered. So, $k = 1$ is the first decimal place of precision; $k = 2$ is the second decimal place; . . . ; $k = \left| K \right|$ is the $\left| K \right|th$ decimal place.  The cardinality of the set K is the precision with which a number, $x_{k}$, is measured. 

Take as examples $x_{k} = 0.478$; and $y_{k} = 0.472$. In these cases, $\left| K \right| = 3$. Start from the first decimal position.
For $k = 1$, we find $x_{k} = y_{k} = 4$. For $k = 2$, $x_{k} = y_{k}$ . But for $k = 3$,  $x_{k} \neq y_{k}$.

We now introduce the following distance (case of vectors $x$ and $y$, with 1 attribute, hence unidimensional):

\begin{equation}
\label{eq:baire}
\db(x_{K}, y_{K}) =   
	\left\{ 
	\begin{array}{ll}
       1 &\;\; $if$\;\;  x_{1} \neq y_{1}\\
       $inf$\;\;  2^{-k} & \;\;\;\;\;\; x_{k} = y_{k} 
                        \;\;\; 1 \leq k \leq
       \left| K \right|
    \end{array}
    \right.
\end{equation}
We call this $\db$ value Baire distance, which is seen to be 
an 
ultrametric~\cite{Murtagh04, Murtagh04-2, Murtagh04-1, Murtagh05, murtagh08} 
distance.

Note that the base 2 is given for convenience. When dealing with binary 
data $x, y$, then 2 is the chosen base. When working with real numbers 
the base can be redefined to 10 if needed.

\subsection{Constructive Hierarchical Clustering Algorithm versus Hierarchical
Encoding of Data}

The Baire distance was introduced and described by Bradley \cite{bradley} 
in the context of inducing a hierarchy on strings over finite alphabets.  
This work further pursued the goal of embedding a dendrogram in a 
p-adic Bruhat-Tits tree, informally characterized as a ``universal 
dendrogram''.  

By convention we denote a prime by p, and a more general, prime or 
non-prime, positive integer by m.  

A geometric foundation for ultrametric structures is presented in 
Bradley \cite{bradley0}.   Starting from the point of view that a 
dendrogram, or ranked or unranked, binary or more general $m$-way, tree,
is an object in a p-adic geometry, it is noted that:  
``The consequence of using p-adic methods
is the shift of focus from imposing a hierarchic structure on data to 
finding a p-adic encoding which reveals the inherent hierarchies.''

This summarizes well our aim in this work.  We seek hierarchy and 
rather than using an agglomerative hierarchical clustering algorithm 
which is of quadratic computational time (i.e., for $n$ individuals or 
observation vectors, $O(n^2)$ computational time is required) we instead
seek to read off a p-adic or m-adic tree.  In terms of a tree, p-adic 
or m-adic mean p-way or m-way, respectively, or that each node in the tree
has at most p or m, respectively, sub-nodes.  

Furthermore, 
by ``reading off'' we are targeting a linear time, or $O(n)$ algorithm
involving one scan over the dataset, and we are imposing thereby 
an encoding of the data.  (We recall that $n$ is the number of 
observations, or cardinality of the observation set $X$.) 

In practice we will be more interested in this work in the hierarchy,
and the encoding algorithm used is a means towards this end.  For a
focus on the encoding task, see \cite{steklov}.

\section{The Set of Clusters Perspective}
\label{section:clusters}

\subsection{The Baire Ultrametric as a Generalized Ultrametric}
\label{genum}

While the Baire distance is also an ultrametric, it is interesting
to note some links with other closely related data analysis and 
computational 
methods.  We can, for example, show a relationship between the Baire
distance and the generalized ultrametric, which maps the cross-product
of a set with itself into the power set of that set's attributes.  
A (standard) ultrametric instead maps the cross-product
of a set with itself into the non-negative reals.  
We pursue this link with the generalized ultrametric in section
\ref{sect231}.  

We also discuss the data analysis method known as Formal Concept 
Analysis as a special case of generalized ultrametrics.  This 
is an innovative vantage point on Formal Concept Analysis because
it is usually motivated and described in terms of lattices, which 
structure the data to be analyzed.  
We pursue this link with Formal Concept Analysis in section \ref{sect232}.  

We note that agglomerative hierarchical clustering, expressed as a 
2-way (or ``binary'') tree, has been related to lattices by, e.g.,
Lerman \cite{Lerman81}, Janowitz \cite{jan2010}, and others.  

\subsubsection{Generalized Ultrametrics}
\label{sect231}

In this section, our focus is on the clusters determined, and on the 
relationships between them.  What we pursue is exemplified as follows. 
Take $x = 0.4578, y = 0.4538$.  Consider the Baire distance between 
$x$ and $y$ as (base 10) $10^{-2}$.   Let us look at the cluster where 
they share membership -- it is the cluster defined by common first 
digit precision and common second digit precision.  We are interested
in a set of such clusters in this section.

The usual ultrametric is an ultrametric distance, i.e.\ for a set I, $d: I \times I \longrightarrow \mathbb{R}^+$.  Thus, the ultrametetric distance is 
a positive real.

The generalized ultrametric is also consistent with this definition, where the range is a subset of the power set: $d: I \times I \longrightarrow \Gamma$, where $\Gamma$ is a partially ordered set with least element.  See \cite{seda}. 
The least
element is a generalized way of seeing zero distance.  
Some areas of application of generalized ultrametrics will now be discussed.

Among other fields, generalized ultrametrics are used in reasoning.  
In the theory of reasoning, a  monotonic operator is rigorous application of a succession of  conditionals (sometimes called consequence relations).  However negation or multiple valued logic (i.e.\ encompassing intermediate truth and falsehood) requires support for non-monotonic reasoning, where fixed
points are modeled as tree structures.  See \cite{seda}. 

A direct application of generalized ultrametrics to data mining is the following. The potentially huge advantage of the generalized ultrametric is that it allows a hierarchy to be read directly off the $I \times J$ input data, and bypasses the $O(n^2)$ consideration of all pairwise distances in agglomerative hierarchical clustering.  Let us assume that the hierarchy is induced on the 
observation set, $I$, which are typically given by the rows of the input 
data matrix.  In \cite{murtagh08} we study application to chemoinformatics.  Proximity and best match finding is an essential operation in this field.  Typically we have one million chemicals upwards, characterised by an approximate 1000-valued attribute encoding.  The set of attributes is $J$, and the number of attributes is the 
cardinality of this set, $| J |$.  

Consider first our need to normalize the data.  We divide each 
boolean (presence/absence) attribute value by its
corresponding column sum.

We can consider the hierarchical cluster analysis from abstract posets as based on a distance or even dissimilarity $d: I \times I 
\rightarrow \mathbb{R}^{|J|}$. The $|J|$-dimensional reals are the domain here. 

As noted in section \ref{intro}, we can consider embedded clusters 
corresponding to the minimal Baire distance (in definition (\ref{eq:baire})
this is seen to be $2^{-1} = 0.5$).  The Baire distance induces 
the hierarchical clustering, and this hierarchical clustering is determined
from the Baire disances.  So it is seen how the Baire distance maps 
onto real valued numbers (cf.\ definition (\ref{eq:baire})) and as such is a
metric.  But the Baire distance also maps onto a hierarchical clustering, 
i.e.\ a partially ordered set of clusters, and so, in carrying out this 
mapping, the Baire distance gives rise to a generalized ultrametric.  

Our Baire-based distance and simultaneously ultrametric is a particular 
case of the generalized ultrametric.  

Figures~\ref{fig:baire-4-3} and \ref{fig:baire-2-1}, to be studied below in section \ref{subsection:clustering-based-baire},  
show how a set of results, related to the range set, $\mathbb{R}^{|J|}$, which are -- in practice -- further processed in order to provide the cluster memberships.  

\subsubsection{Link with Formal Concept Analysis}
\label{sect232}

We pursue the case of an ultrametric defined on the power set or join semilattice.  Comprehensive background on ordered sets and lattices can be found in \cite{davey}. A review of generalized distances and ultrametrics can be found in \cite{sedacj}.

Typically hierarchical clustering is based on a distance (which can be relaxed often to a dissimilarity, not respecting the triangular inequality, and {\em mutatis mutandis} to a similarity), defined on all pairs of the object set: $d: X \times X \rightarrow \mathbb{R}^{+}$.  I.e., a distance is  a positive real value.  Usually we require that a distance cannot be 0-valued unless the objects are identical.  That is the traditional approach.

A different form of ultrametrisation is achieved from a dissimilarity defined on the power set of attributes characterising the observations (objects, individuals, etc.) $X$.  Here we have: $d : X \times X \longrightarrow 2^J$, where $J$ indexes  the attribute (variables, characteristics, properties, etc.) set. 

This gives rise to a different notion of distance, that maps pairs of objects onto elements of a join semilattice.  The latter can represent all subsets of the attribute set, $J$.  That is to say, it can represent the power set, commonly denoted $2^J$, of $J$.

As an example, consider, say, $n = 5$ objects characterised by 3 boolean (presence/absence) attributes, shown in Figure \ref{figfca} (top). Define dissimilarity between a pair of objects in this table as a {\em set} of 3 components, corresponding to the 3 attributes, such that if both components are 0, we have 1; if either component is 1 and the other 0, we have 1; and if both components are 1 we get 0.  This is the simple matching coefficient.  
We could use, e.g., Euclidean distance for each of the values sought; but here instead we treat 0 values in both components as signalling  a 1 contribution (hence, 0 is a data encoding of a property rather than its absence).  We get then $d(a,b) = 1, 1, 0$ which we will call \verb+d1,d2+.  Then, $d(a,c) = 0, 1, 0$ which we will call \verb+d2+.  Etc. With the latter, \verb+d1,d2+ here, \verb+d2+, and so on, we create lattice nodes as shown in the middle part of Figure \ref{figfca}.
So, note in this figure, how the order relation holds between \verb+d1,d2+ at 
level 2 and \verb+d2+ at level 1.  

\begin{figure}
\begin{center}
\begin{tabular}{cccc}
   &  $v_1$  &   $v_2$  & $v_3$  \\
a  &    1    &    0     &   1    \\
b  &    0    &    1     &   1    \\
c  &    1    &    0     &   1    \\
e  &    1    &    0     &   0    \\
f  &    0    &    0     &   1    \\
\end{tabular}
\end{center}
\begin{verbatim}                                                                    
                                                                                    
Potential lattice vertices      Lattice vertices found       Level                  
                                                                                    
       d1,d2,d3                        d1,d2,d3                3                    
                                         /  \                                       
                                        /    \                                      
  d1,d2   d2,d3   d1,d3            d1,d2     d2,d3             2                    
                                        \    /                                      
                                         \  /                                       
   d1      d2      d3                     d2                   1                    
                                                                                    
\end{verbatim}

   The set d1,d2,d3 corresponds to:     $d(b,e)$ and $d(e,f)$

   The subset d1,d2 corresponds to:     $d(a,b), d(a,f), d(b,c),                    
                                        d(b,f),$ and $d(c,f)$

   The subset d2,d3 corresponds to:     $d(a,e)$ and $d(c,e)$

   The subset d2 corresponds to:        $d(a,c)$

\medskip

   Clusters defined by all pairwise linkage at level $\leq  2$:

$   a, b, c, f$

$   a, c, e$

\medskip

   Clusters defined by all pairwise linkage at level $\leq 3$:

$   a, b, c, e, f$

\caption{Top: example data set consisting of 5 objects,
characterized by 3 boolean attributes. Then: lattice corresponding to this data and its interpretation.}
\label{figfca}
\end{figure}

In Formal Concept Analysis \cite{davey,ganter}, it is the lattice itself which is of primary interest.  In \cite{jan2010} there is discussion of, and a range of examples on,  the close relationship between the traditional hierarchical cluster analysis based on $d: I \times I \rightarrow \mathbb{R}^{+}$, and hierarchical cluster analysis ``based on abstract posets'' (a poset is a partially ordered set), based on $d: I \times I \rightarrow 2^{J}$.  The latter, leading to clustering based on dissimilarities, was developed initially in \cite{jan0}.

Thus, in Figure \ref{figfca}, we have $d(a,b) \rightarrow$ \verb+d1,d2+,
$(a,f) \rightarrow$ \verb+d1,d2+, $d(a,e) \rightarrow$ \verb+d2,d3+, 
and so on.  We note how the \verb+d1,d2+ etc.\ are sets that are 
subsets of the power set of attributes, $2^J$.   

\section{A Baire-Based Hierarchical Clustering Algorithm}
\label{subsection:alg}

We have discussed Formal Concept Analysis as a particular case of 
the use of generalized ultrametrics.  We noted that a nice feature of the
generalized ultrametric is that it may allow us to directly ``read off'' a
hierarchy.   That in turn, depending of course on the preprocessing steps 
needed or other properties of the algorithm, may be computationally very 
efficient. 

Furthermore, returning further back to section \ref{subsection:baire}, we 
note that the ultrametric Baire space can be viewed in a generalized ultrametric
way.  We can view the output mapping as being a restricted subset of the 
power set of the set $K$ of digits of precision.   Alternatively expressed,
the output mapping is a restricted subset of the power set, $2^K$.   Why 
restricted? -- because we are only interested in a longest common prefix 
sequence of identical digits, and not in the sharing of any arbitrary 
precision digits.   

A straightforward algorithm for hierarchical clustering based on the Baire
distance, as described in section \ref{subsection:baire} is as follows. 
Because of working with real numbers in our case study below, we define 
the base in relation (\ref{eq:baire}) as 10 rather than 2.  

For the first digit of precision, $k = 1$, consider 10 ``bins'' corresponding 
to the digits $0, 1, \dots , 9$.  For each of the nodes corresponding to these
bins, consider 10 subnode bins corresponding to the second digit of precision, 
$k = 2$, associated with $0, 1, \dots , 9$ at this second level.  We can 
continue for a third and further levels.  In practice we will neither 
permit nor wish for a very deep (i.e., with many levels) storage tree.  For the
base 10 case, it is convenient for level one (corresponding to $k = 1$) to give rise to up to 10 
clusters.   For level two (corresponding to $k = 2$) we have up to 100 
clusters.  We see that in practice a small number of levels will suffice.  
In one pass over the data we map each observation (recall that it is univariate
but we are using its ordered set of digits, i.e.\ ordered set $K$) to its 
bin or cluster at each level.   For $\ell$ levels, the computation required
is $n \cdot \ell$ operations.   For a given value of $\ell$ we therefore 
have $O(n)$ computation -- and furthermore with a very small constant of 
proportionality since we are just reading off the relevant digit and,
presumably, updating a node or cluster membership list and cardinality.  

\section{Astronomical Case Study}

\subsection{The Sloan Digital Sky Survey}
The Sloan Digital Sky Survey (SDSS)~\cite{SDSS} is systematically mapping the sky producing a detailed image of it and determining the positions and absolute brightnesses of more than 100 million celestial objects. It is also measuring the distance to a million of the nearest galaxies and to a hundred thousand quasars.  The acquired data has been openly given to the scientific community.

\begin{figure}[ht!]
  \begin{center}
   \includegraphics[scale=0.45]{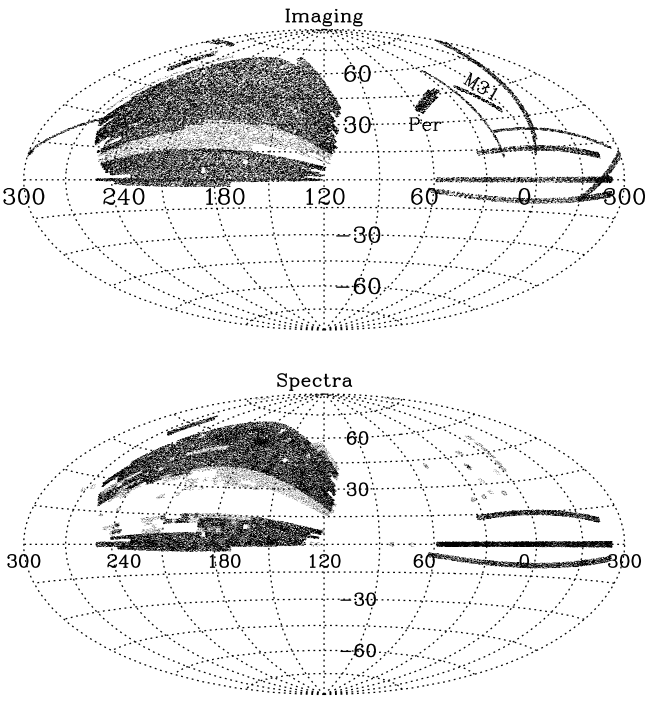}
    \caption{Distribution in the sky of the SDSS Data Release 5~\cite{Adelman-McCarthy07}.}
  \label{fig:sdss-dr5}
  \end{center}
\end{figure}

Figure~\ref{fig:sdss-dr5} depicts the SDSS Data Release 5 for imaging and spectral data. For every object a large number of attributes and measurements are acquired.
See ~\cite{Adelman-McCarthy07} for a description of the data available in this catalog.

In particular we use the data that has been studied by 
Longo group~\cite{Longo10} and used intensively by Longo and 
D'Abrusco~\cite{Dabrusco07-1,Dabrusco07-2,Dabrusco06-1}.

\subsection{Doppler Effect and Redshift}

Light from moving objects will appear to have different wavelengths depending on the relative motion of the source and the observer. On the one hand we have that if an object is moving towards an observer, the light waves will be compressed from the observer viewpoint, then the light will be shifted to a shorter wavelength or it will appear to be blue shifted. On the other hand if the object is moving away from the observer, the light wavelength will be expanding, thus red shifted. This is also called Doppler effect (or Doppler shift) named after the Austrian physicist Christian Doppler, who first described this phenomenon in 1845.
A very important piece of information obtained in cosmology from the Doppler shift is to know if an object is moving towards or away from us, and the speed at which this is happening.

\textbf{Spectrometric measurement of redshift}: under certain conditions all atoms can be made to emit light, doing so at particular wavelengths, which can be measured accurately.  Chemical compounds are a combination of different atoms working together. Thus, when measuring the precise wavelength at which a particular chemical radiates we are effectively obtaining a signature of this chemical. These emissions are seen as lines (emission or absorption) in the electromagnetic spectrum. For example, hydrogen is the simplest chemical element with atomic number 1, and also is the most abundant chemical in the universe. Hydrogen has emission lines at 6562.8 \AA, 4861.3 \AA, 4340 \AA, 4102.8 \AA, 4102.8 \AA, 3888.7 \AA, 3834.7 \AA ~and 3798.6 \AA ~(where~\AA~is an Angstrom equal to $10^{-10}$m). If the spectrum of a celestial body has emission lines in these wavelengths we can conclude that hydrogen is present there.
 
\textbf{Photometric measurement of redshift}: sometimes obtaining spectrometric measurements can be very difficult due to the large number of objects to observe or because the signal is too weak for the current spectrometric techniques. A redshift estimate can be obtained using large/medium band photometry instrumentation instead of spectrometric. This technique is based on the identification of strong spectral features. This is much faster than spectrometric measurement but also of lesser quality and precision~\cite{Fernandez01}. 

Hence the context of our clustering work is to see how well the more easily obtained photometric redshifts can be used as estimates for the spectrometric redshifts that are obtained with greater cost.  We limit our work here to the fast finding of clusters of associated photometric and spectrometric redshifts.   In doing so, we find some interesting new ways of finding good quality mappings from photometric to spectrometric redshifts with high confidence.

\section{Inducing a Hierarchy on the SDSS Data using the Baire Ultrametric}

The aim here is to build a mapping from $z_{spec} \rightarrow z_{phot}$ to 
help calibrating the redshifts, based on the $z_{spec}$ observed values. 
Traditionally we could map $ f : z_{phot} \longrightarrow z_{spec}$  based 
on trained data.  That is to say, having set up the calibration, we determine
the higher quality information from the more readily available less high 
quality information.   
The mapping $f$ could be linear (e.g.\ linear regression) or non-linear (e.g.\ multilayer perceptron) as used by D'Abrusco~\cite{Dabrusco06-1}. These techniques are  global. Here our interest is to develop a locally adaptive approach based on numerical precision.  That is the direct benefit of the (very fast, hierarchical) 
clustering based on the Baire distance.

We look specifically into four parameters: right ascension (RA), declination (DEC), spectrometric ($z_{spec}$) and photometric ($z_{phot}$) redshift. Table~\ref{tab:sdss-data-format} shows a small subset of the data used for experimentation and analysis. 

As already noted the spectrometric technique uses the spectrum of electromagnetic radiation (including visible light) which radiates from stars and other celestial objects.  The photometric technique uses a faster and economical way of measuring the redshifts. 
\bigskip\bigskip

\begin{table}[ht!]
\centering
\scalebox{0.9}{
\begin{tabular}{|l|l|l|l|} \hline 
	\hfill RA & \hfill DEC & \hfill Spec & \hfill Phot \\\hline\hline
	145.4339  & 0.56416792 & 0.14611299	 & 0.15175095  \\\hline 
	145.42139 & 0.53370196 & 0.145909	 & 0.17476539  \\\hline
	145.6607  & 0.63385916 & 0.46691701	 & 0.41157582  \\\hline
	145.64568 & 0.50961215 & 0.15610801	 & 0.18679948  \\\hline 
\end{tabular}} 
\caption{Data format for right ascension, declination, $z_{Spec}$ and $z_{Phot}$.} 
\label{tab:sdss-data-format}
\end{table}

\subsection{Clustering SDSS Data}
\label{subsection:clustering-based-baire}

We use clustering to support a  nearest neighbor regression.  Hence we are 
interested in the matching up to some level of precision between pairs of 
$z_{spec}$ and $z_{phot}$ values that are assigned to the same cluster.  

In order to perform the clustering process introduced in 
section~\ref{subsection:baire} and further described in \ref{subsection:alg}, 
we compare every $z_{spec}$ and $z_{phot}$ data point searching for common 
prefixes based on the longest common prefix (see section~\ref{subsection:baire}). Thereafter, the data points that have digit coincidences are grouped together to form clusters.

\begin{figure}[ht!]
   \includegraphics[scale=.65]{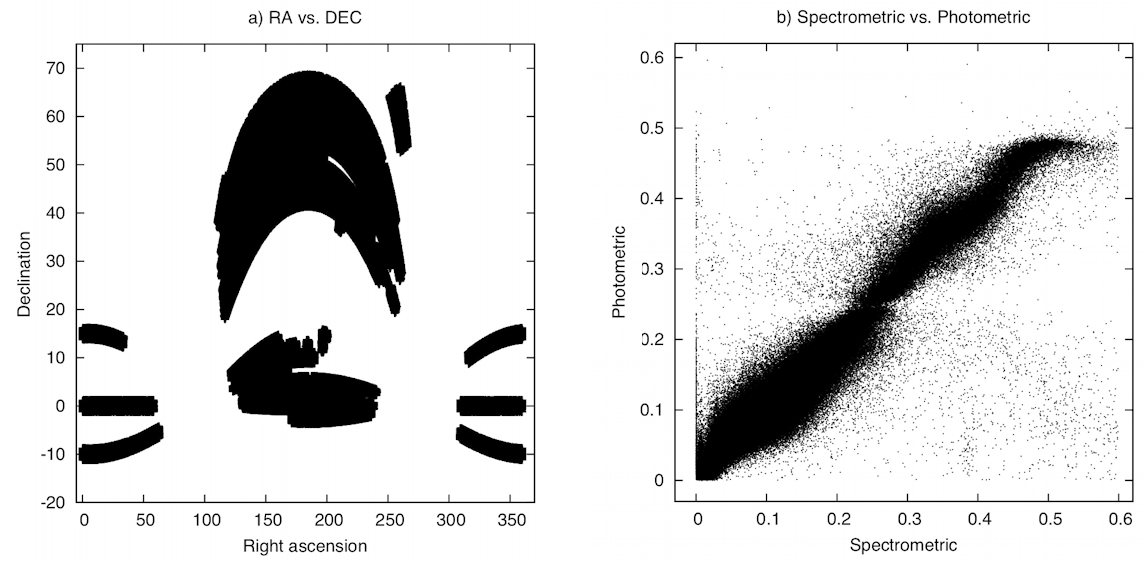}
    \caption{Left: right ascension (RA) versus declination (DEC); Right: $z_{spec}$ versus $z_{phot}$. SDSS data selection used for redshift analysis.}
  \label{fig:characterisation}
\end{figure}

\begin{figure}[h!]
\centering
	\includegraphics[scale=.65]{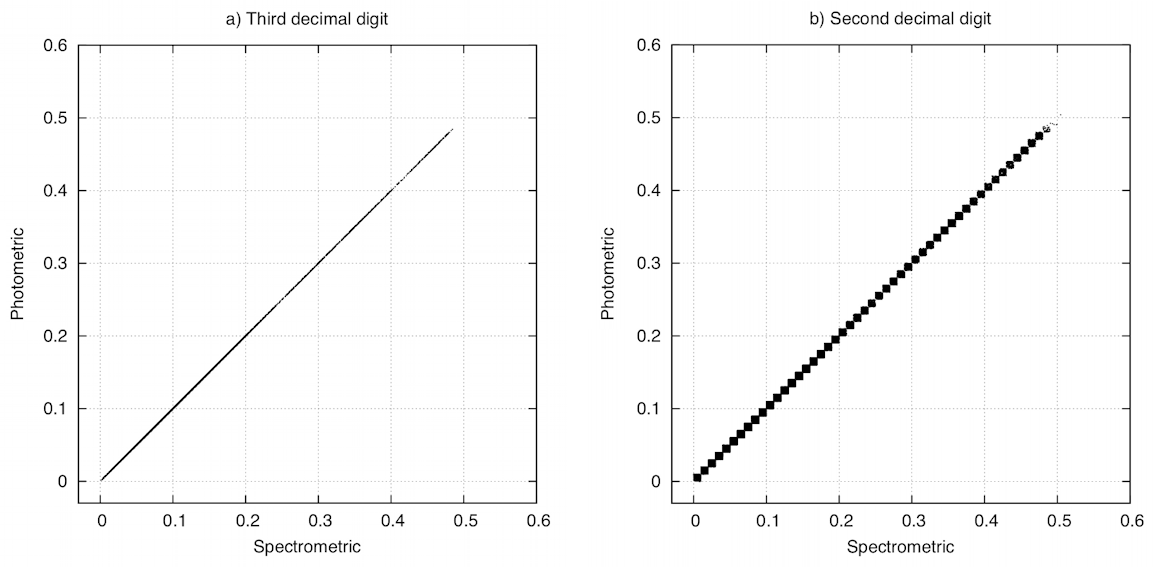}
    \caption{Prefix-wise clustering frequencies depicting 3rd decimal digit coincidences (left panel), and two decimal digit coincidences (right panel).}
	\label{fig:baire-4-3}
\end{figure}

\begin{figure}[h!]
\centering   
	\includegraphics[scale=.65]{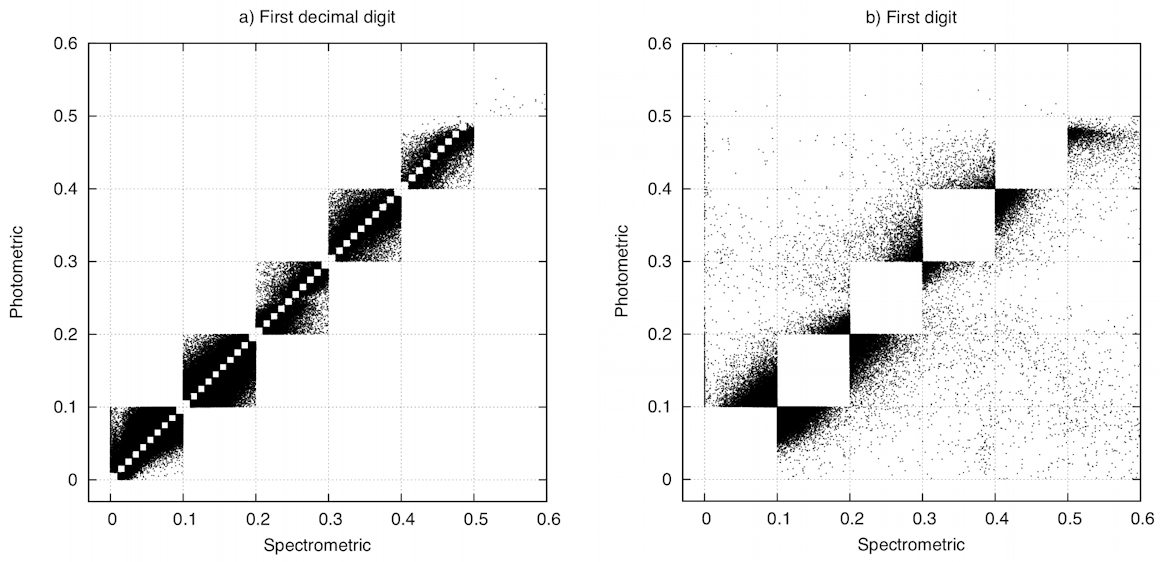}
    \caption{Prefix-wise clustering frequencies depicting the 1st decimal digit coincidences (left panel), and first digit coincidences (right panel).}
  \label{fig:baire-2-1}
  
\end{figure}

Data characterisation is presented in Figure~\ref{fig:characterisation}. The left panel shows the $z_{spec}$ and $z_{phot}$ sky coordinates of the data currently used by us to cluster redshifts. This section of the sky presents approximately 0.5 million object coordinate points with the current data. As can be observed, various sections of the sky are represented in the data. We find this useful since preliminary data exploration has shown that correlation between $z_{spec}$ and $z_{phot}$ is consistent in different parts of the sky. 
For example, when taking correspondences between $z_{spec}$ and $z_{phot}$ as shown in Figures~\ref{fig:baire-4-3} and~\ref{fig:baire-2-1}, and plotting them in RA and DEC space (i.e. astronomical coordinate space) we have the same shape as presented in Figure~\ref{fig:characterisation}. 

This leads us to conclude  that digit coincidences of the redshift measures are distributed approximately uniformly in the sky and are not concentrated spatially. The same occurs for all the other clusters.
We will concentrate on the very near astronomical objects, represented by redshifts between $0$ and $0.6$.  When we plot $z_{spec}$ versus $z_{phot}$ we obtain a highly correlated signal as shown in Figure~\ref{fig:characterisation}, right panel. The number of observations that we therefore analyse is 443,014.

\begin{figure}[hb!]
  \begin{center}
   \includegraphics[scale=.7]{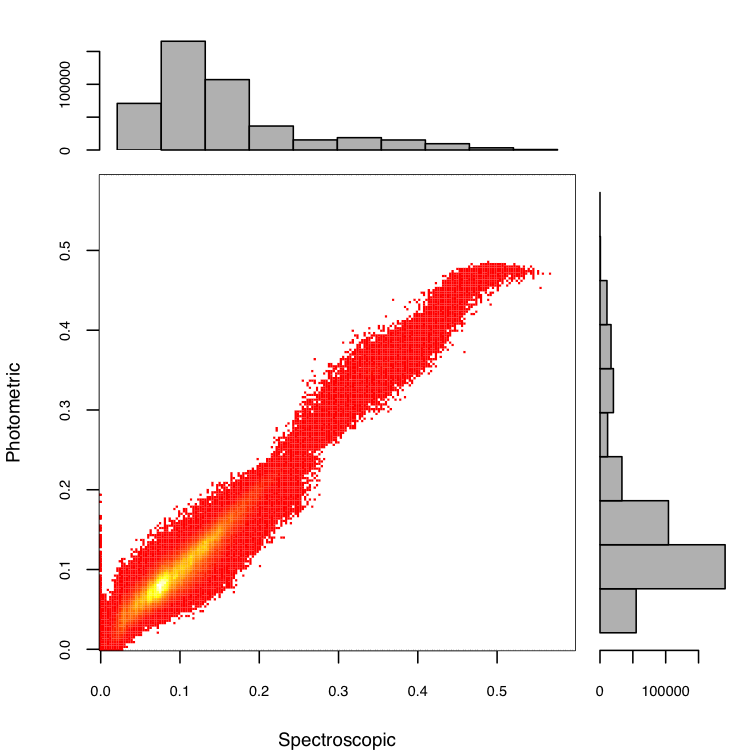}
    \caption{Heat plot and histogram for $z_{spec}$ versus $z_{phot}$. Histogram at the top shows the $z_{spec}$ frequencies, histogram at the right shows $z_{phot}$ frequencies.}
  \label{fig:heat}
  \end{center}
\end{figure}	
\clearpage

Looking at Figure~\ref{fig:heat} it can be seen clearly that most data points fall in the range between 0 and 0.2. Here the histogram on the top shows the $z_{phot}$ data points distribution, and the histogram on the right the $z_{spec}$ data points distribution.  The heat plot also highlights the area where data points are concentrated, where the yellow colour (white region in monochrome print) shows the major density. 

Consequently, now we know that most cluster data points will fall within this range (0 and 0.2) if common prefixes of digits in the redshift values, taken as strings, are found.   

Figures~\ref{fig:baire-4-3} and~\ref{fig:baire-2-1} show graphically how $z_{spec}$ and $z_{phot}$ correspondences look at different levels of decimal precision. On one hand we find that values of $z_{spec}$ and $z_{phot}$ that have equal precision up to the 3rd decimal digit are highly correlated. On the other hand when $z_{spec}$ and $z_{phot}$ have only the first digit in common, correlation is weak.
For example, let us consider the following situations for plots~\ref{fig:baire-4-3} and~\ref{fig:baire-2-1}:

\begin{itemize}
	\item Figure~\ref{fig:baire-4-3} left: let us take the values of  $z_{spec} = 0.437$ and $z_{phot}= 0.437$. We have that they share the first digit, the first decimal digit, the second decimal digit, and the third decimal digit. Thus, we have a highly correlated signal of the data points that share only up to the third decimal digit.
	\item Figure~\ref{fig:baire-4-3} right: let us take the values of  $z_{spec} = 0.437$ and $z_{phot}= 0.439$. We have that they share the first digit, the first decimal digit, and the second decimal digit. Therefore, the plot shows data points that share only up to the second decimal digit.
	\item Figure~\ref{fig:baire-2-1} left: let us take the values of  $z_{spec} = 0.437$ and $z_{phot}= 0.474$. We have that they share the first digit, and the first decimal digit. Thus, the plot shows data points that share only up to the first decimal digit.
	\item Figure~\ref{fig:baire-2-1} right: let us take the values of  $z_{spec} = 0.437$ and $z_{phot}= 0.571$. We have that they share only the integer part of the value, and that alone. Furthermore, this implies redshifts that do not match in succession of decimal digits.	 For example, if we take the values 0.437 and 0.577, the fact that the third digit is 7 in each case is not of use. 
\end{itemize}

\clearpage

\begin{figure}[h!]
  \begin{center}
   \includegraphics[scale=.58]{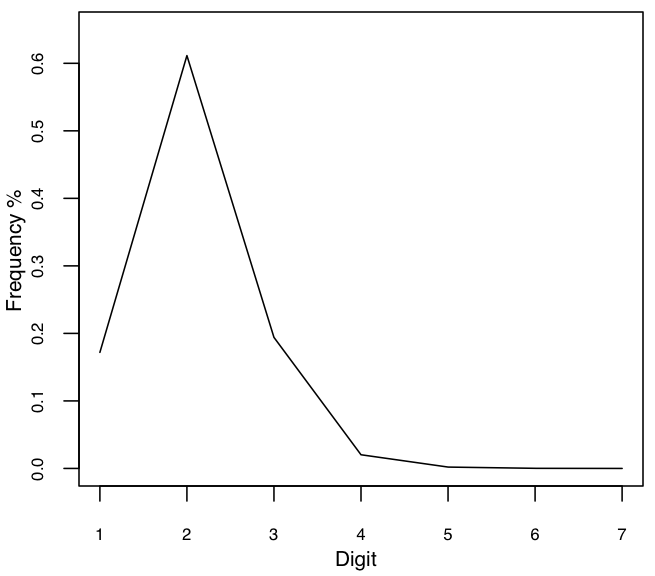}
    \caption{Frequency distribution for Table~\ref{tab:prefixwiseClustering}. The abscissa shows the digit positions, where 1 is the first digit, 2 the first decimal digit, 3 the second decimal digit and so on.}
  \label{fig:freq}
  \end{center}
\end{figure}

Table~\ref{tab:prefixwiseClustering} (see also Figure \ref{fig:freq}) 
shows the clusters found for all different levels of precision. In other words this table allows us to define empirically the confidence levels for mapping of  $z_{phot}$ and $z_{spec}$. For example, we can expect that $82.8\%$ of values for $z_{spec}$ and $z_{phot}$ have at least two common prefix digits. This percentage of confidence is derived as follows: the data points that share six, five, four, three, two, and one decimal digit (i.e., $4 + 90 + 912 + 8,982 + 85,999 + 270,920 = 366,907$ data points. Therefore 82.8\% of the data). 
Additionally we observe that around a fifth of the observations share at least 3 digits in common. Namely, $4 + 90 + 912 + 8,982 + 85,999  = 95,987$ data points, which equals $21.7\%$ of the data.

\begin{table}[ht!]
\centering
\scalebox{0.9}{
\begin{tabular}{|c|r|c|}
	\hline Digit          & No. \hspace{5pt}& \%     \\\hline
	\hline   1	          &  76,187	        & 17.19  \\\hline 
	\hline  Decimal digit & No. \hspace{5pt}& \%     \\\hline
	\hline   1	          & 270,920      	& 61.14  \\\hline 
	         2	          &  85,999	        & 19.40  \\\hline 
	         3	          &   8,982         	&  2.07  \\\hline 
	         4	          &     912      	&  0.20  \\\hline
	         5	          &      90      	&  0.02  \\\hline 
	         6	          &       4	        &  ---   \\\hline
	  \hline              &  443,094        & 100    \\\hline 
\end{tabular}} \bigskip
\caption{Data points based on the longest common prefix for different levels of precision. This includes the integer part of a data point (first digit) and the decimal digits of a data point (first to sixth digit).} 
\label{tab:prefixwiseClustering}
\end{table}

\clearpage

\section{Comparative Evaluation with $k$-Means}

In this section we 
compare the Baire-based clustering to results obtained with the widely-used 
$k$-means clustering algorithm.

\subsection{Baire-Based Clustering and $k$-Means Cluster Comparison}

In order to establish how ``good'' the Baire clusters are we can compare them with clusters resulting from the $k$-means algorithm. Let us recall that our data values are in the interval $\left[0, 0.6\right[$ (i.e.\ including zero values but excluding 0.6).  Additionally, we have seen that the Baire distance is an ultrametric that is strictly defined in a tree. Thus, when building the Baire based clusters we will have a root node ``0'' that includes all the observations (every single data point analysed starts with 0).  For the Baire distance with exponent $-2$  we have six nodes (or clusters) with indices ``00, 01, 02, 03, 04, 05''.  For the Baire distance of exponent $-3$ we have 60 clusters with indices ``000, 001, 002, 003, 004,...,059'' (i.e. ten children for each node 00,..,05).   (Cf.\ how this adapts the discussion in 
section \ref{subsection:alg} in a natural way to our data.)  

We carried out a number of comparisons for the Baire distance of two and three. For example, by design we have that for $\db=10^{-2}$ there are six clusters. Thus we took our data set and applied $k$-means with six centroids based on an implementation from the Hartigan and Wong~\cite{Hartigan79} algorithm. Euclidean distance is used, as usual, here.  The results can be seen in Table~\ref{tab:kmeans-6}, where the columns are the $k$-means clusters and the rows are the Baire clusters. From the Baire perspective we see that the node 00 has 97084 data points contained within the first $k$-means cluster and 64950 observations in the fifth. Looking at node 04, all members belong to the third cluster of $k$-means. We can see that the Baire clusters are closely related to the clusters produced by $k$-means at a given level of resolution.
\bigskip\bigskip

\begin{table}[ht!]
\centering
\scalebox{0.9}{
\begin{tabular} {|c|c|c|c|c|c|c|} 
	\hline 	--- 	& 1 & 5 & 4 & 6 & 2 & 3 
	\\\hline
	  \hline \hspace{5pt}00\hspace{5pt} & 97084 & 64950 & 0      & 0     & 0     & 0 	
	\\\hline \hspace{5pt}01\hspace{5pt} & 0     & 28382 & 101433 & 14878 & 0     & 0 
	\\\hline \hspace{5pt}02\hspace{5pt} & 0     & 0     & 0      & 18184 & 4459  & 0 
	\\\hline \hspace{5pt}03\hspace{5pt} & 0     & 0     & 0      & 0     & 25309 & 1132 
	\\\hline \hspace{5pt}04\hspace{5pt} & 0     & 0     & 0      & 0     & 0     & 11116 
	\\\hline \hspace{5pt}05\hspace{5pt} & 0     & 0     & 0      & 0     & 0     & 21
\\\hline 
\end{tabular}}
\caption{Cluster comparison based on $\db = 10^{-2}$. Columns show the $k$-means clusters, and the rows show the Baire clusters. The cells present the number of data points for a given cluster.}
\label{tab:kmeans-6}
\end{table}

\begin{figure}[h!]
  \begin{center}
  \includegraphics[scale=.75]{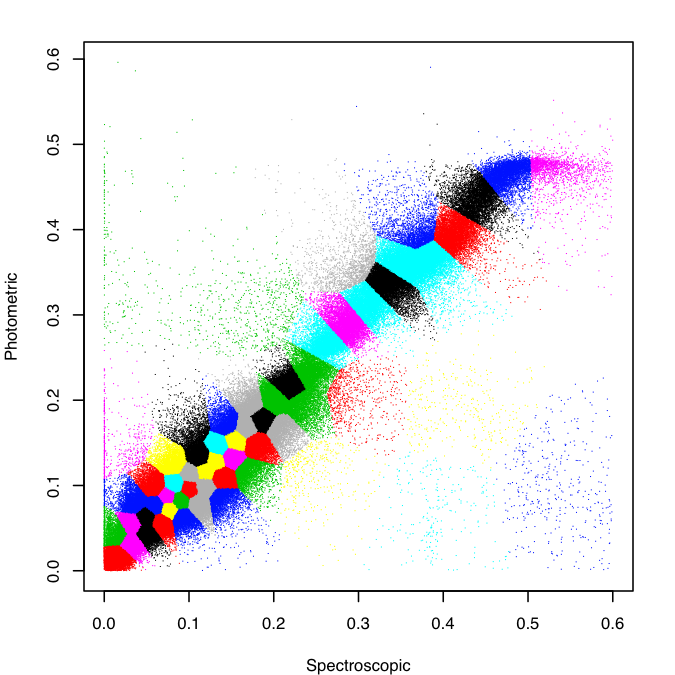}
    \caption{$K$-means clustering for $k=60$ after 38 iterations. Note that non-contiguous groups may be colored the same.}
  \label{fig:kmeans-60}
  \end{center}
\end{figure}

We can take this procedure further and compare the clusters for $\db$ defined from 3 digits of precision, and $k$-means with $k=60$ centroids as observed in Figure~\ref{fig:kmeans-60}. 

Looking at the results from the Baire perspective we find that 27 clusters are overlapping, 9 clusters are empty, and 24 Baire clusters are completely within the boundaries of the ones produced by $k$-means as presented in Table~\ref{tab:kmeans-60}. This last result is better seen in Table~\ref{tab:kmeans-60-short}, which is the subset of Table~\ref{tab:kmeans-60} (see Appendix~\ref{appendixA}) where complete matches are shown. These tables have been row and column permuted in order to clearly appreciate the correspondences.
 
It is seen that the match is consistent even if there are differences due to the different clustering criteria at issue.  We have presented results in such a way as to show both consistency and difference.  

\begin{table}[hb!]
\centering
\scalebox{0.7}{
\begin{tabular}{|c|c|c|c|c|c|c|c|c|c|c|c|c|c|c|c|c|} 
	\hline 	---   &  21 &   1 &   6 &  38 & 25 & 58 & 32 & 20 & 15 & 13 & 14 & 37 & 17 &  2 & 51 &  4 
	 \\\hline
	  \hline  015 &3733 &   0 &   0 &   0 &  0 &  0 &  0 &  0 &  0 &  0 &  0 &  0 &  0 &  0 &  0 &  0 
	 \\\hline 004 &   0 &3495 &   0 &   0 &  0 &  0 &  0 &  0 &  0 &  0 &  0 &  0 &  0 &  0 &  0 &  0
	 \\\hline 018 &   0 &   0 &2161 &   0 &  0 &  0 &  0 &  0 &  0 &  0 &  0 &  0 &  0 &  0 &  0 &  0
	 \\\hline 020 &   0 &   0 &   0 &1370 &  0 &  0 &  0 &  0 &  0 &  0 &  0 &  0 &  0 &  0 &  0 &  0
	 \\\hline 001 &   0 &   0 &   0 &   0 &968 &  0 &  0 &  0 &  0 &  0 &  0 &  0 &  0 &  0 &  0 &  0
	 \\\hline 000 &   0 &   0 &   0 &   0 &515 &  0 &  0 &  0 &  0 &  0 &  0 &  0 &  0 &  0 &  0 &  0
	 \\\hline 022 &   0 &   0 &   0 &   0 &  0 &896 &  0 &  0 &  0 &  0 &  0 &  0 &  0 &  0 &  0 &  0
	 \\\hline 034 &   0 &   0 &   0 &   0 &  0 &  0 &764 &  0 &  0 &  0 &  0 &  0 &  0 &  0 &  0 &  0
	 \\\hline 036 &   0 &   0 &   0 &   0 &  0 &  0 &  0 &652 &  0 &  0 &  0 &  0 &  0 &  0 &  0 &  0
	 \\\hline 037 &   0 &   0 &   0 &   0 &  0 &  0 &  0 &508 &  0 &  0 &  0 &  0 &  0 &  0 &  0 &  0
	 \\\hline 026 &   0 &   0 &   0 &   0 &  0 &  0 &  0 &  0 &555 &  0 &  0 &  0 &  0 &  0 &  0 &  0
	 \\\hline 027 &   0 &   0 &   0 &   0 &  0 &  0 &  0 &  0 &464 &  0 &  0 &  0 &  0 &  0 &  0 &  0
	 \\\hline 032 &   0 &   0 &   0 &   0 &  0 &  0 &  0 &  0 &  0 &484 &  0 &  0 &  0 &  0 &  0 &  0
	 \\\hline 030 &   0 &   0 &   0 &   0 &  0 &  0 &  0 &  0 &  0 &  0 &430 &  0 &  0 &  0 &  0 &  0
	 \\\hline 045 &   0 &   0 &   0 &   0 &  0 &  0 &  0 &  0 &  0 &  0 &  0 &398 &  0 &  0 &  0 &  0
	 \\\hline 044 &   0 &   0 &   0 &   0 &  0 &  0 &  0 &  0 &  0 &  0 &  0 &295 &  0 &  0 &  0 &  0
	 \\\hline 039 &   0 &   0 &   0 &   0 &  0 &  0 &  0 &  0 &  0 &  0 &  0 &  0 &278 &  0 &  0 &  0
 	 \\\hline 024 &   0 &   0 &   0 &   0 &  0 &  0 &  0 &  0 &  0 &  0 &  0 &  0 &  0 &260 &  0 &  0
	 \\\hline 041 &   0 &   0 &   0 &   0 &  0 &  0 &  0 &  0 &  0 &  0 &  0 &  0 &  0 &  0 &231 &  0
	 \\\hline 042 &   0 &   0 &   0 &   0 &  0 &  0 &  0 &  0 &  0 &  0 &  0 &  0 &  0 &  0 &225 &  0
	 \\\hline 047 &   0 &   0 &   0 &   0 &  0 &  0 &  0 &  0 &  0 &  0 &  0 &  0 &  0 &  0 &  0 &350
	 \\\hline 048 &   0 &   0 &   0 &   0 &  0 &  0 &  0 &  0 &  0 &  0 &  0 &  0 &  0 &  0 &  0 & 57
	 \\\hline 049 &   0 &   0 &   0 &   0 &  0 &  0 &  0 &  0 &  0 &  0 &  0 &  0 &  0 &  0 &  0 &  5
	 \\\hline 050 &   0 &   0 &   0 &   0 &  0 &  0 &  0 &  0 &  0 &  0 &  0 &  0 &  0 &  0 &  0 &  1
\\\hline
\end{tabular}}
\caption{Subset of cluster comparison based on $\db = 10^{-3}$; columns show the $k$-means clusters ($k=60$); rows show Baire nodes.} 
\label{tab:kmeans-60-short}
\end{table}
\clearpage

\subsection{Baire and $k$-Means Clustering Time Comparison}

In order to compare the time performances of the Baire and $k$-means algorithms we took $\db=10^{-3}$ as a basis for the test. Let us remember that for $\db=10^{-3}$ we have potentially 60 clusters for the data in the range $\left[0, 0.6\right[$. Looking at the classification from the hierarchical tree viewpoint we have: one cluster for first level (i.e., the root node or first digit); six clusters for the second level (i.e., first decimal digit or 0, 1, 2, 3, 4, and 5); and ten clusters for the third level or second decimal digit.  To obtain the potential number of clusters we multiply the potential nodes for the first, second and third levels of the tree. That is $1 \cdot 6 \cdot 10 = 60$ clusters.

Therefore for the time comparison we have $\db=10^{-3}$ of 60 clusters, which is the parameter given to $k$-means as initial number of centroids.  The other parameter needed is the number of iterations. For $k$-means we are interested in the average time over many runs. Thus, we use average time over 50 executions for each iteration of 1, 5, 10, 15, 20, 28, 30, 35, and 38. 

\begin{figure}[h!]
  \begin{center}
  \includegraphics[scale=.7]{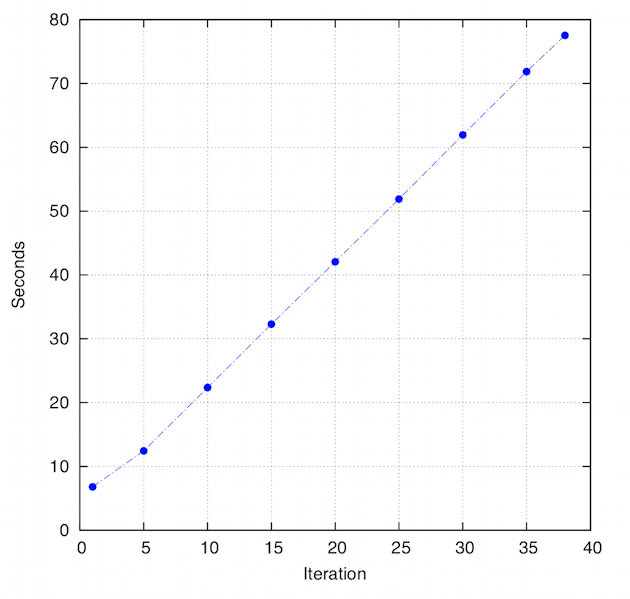}
    \caption{$K$-means average processing time in seconds for $k=60$. Averages are obtained for 9 examples with 50 executions each.}
  \label{fig:kmeans60-times}
  \end{center}
\end{figure}

The results can be observed in Figure~\ref{fig:kmeans60-times}. It is clear that the time in $k$-means is linear with respect to the number of iterations (this is well understood in the $k$-means literature). In this particular case the algorithm converges around the iteration number 38. Note that these executions are based on different random initialisations.  
The times for the $k$-means algorithm were obtained with the R statistical software. These times were faster than the times obtained by the algorithm implemented with Java.  

\begin{table}[hb!]
\centering
\begin{tabular}{|c|c|} \hline 
   Iteration & Average time	\\\hline\hline
	 1		 & 6.81			\\\hline 
	 5		 & 12.44			\\\hline
	10		 & 22.35			\\\hline
	15		 & 32.30			\\\hline 
	20		 & 42.07			\\\hline
	25		 & 51.90			\\\hline
	30		 & 61.94			\\\hline
	35		 & 71.85			\\\hline
	38		 &77.53			\\\hline
\end{tabular} 
\caption{Time average for $k$-means algorithm over 50 executions for each total iteration count.} 
\label{tab:kmeans-time}
\end{table}

The Baire method only needs one pass over the data to produce the clusters. Regarding the time needed, we tested a Java implementation of the Baire algorithm. We ran 50 experiments over the SDSS data. It took on average 2.9 seconds. Compare this to Table~\ref{tab:kmeans-time}.

We recall that this happens because of the large number of iterations involved in the case of $k$-means. Even in the case when just one iteration is considered for $k$-means (note that the algorithm does not converge in that case) the time taken is more than double when compared with the Baire (6.8 seconds versus 2.9 seconds).

\section{Spectrometric and Photometric Digit Distribution}\label{section:digit-distribution}

We have seen that the Baire ultrametric produces a strict hierarchical classification. In the case of $z_{spec}$ and $z_{phot}$ this can be seen as follows. Let us take any observed measurement of either case of $z_{spec} = z_{phot}$. Let us say $z_{spec} = z_{phot} = 0.1257$. Here we have that for $|K|=4$,  $z_{spec} = z_{phot}$. Hierarchically speaking we have that the root node is 0, for the first level where there potentially exist 6 nodes (i.e. 0,1,...,5); for the second level potentially there are 60 nodes; and so on until $k = |K| = 4$, and $z_{spec} = z_{phot}$, where potentially there are $6\cdot 10 \cdot 10 \cdot 10 = 6,000$ nodes. 

Of course not all nodes will be populated.  In fact we can expect that a large number of these potential nodes will be empty if the number of observations $n$ is lower than the potential number of nodes for a certain precision $|K|$ (i.e.\
 $n \leq 10^{|K|}$).  Note that this points to a big storage cost,  but in practice the tree is very sparsely populated and $|K|$ small.

A particular interpretation can be given in the case of an observed data point. Following up the above example if we take  $z_{spec} = z_{phot} = 0.1257$, a tree can be produced to store all observed data that falls within this node. Doing this has many advantages from the viewpoint of storing. Access and retrieval, for example, is very fast and it is easy to retrieve all the observations that fall within a given node and its children.

\begin{figure}[h!]
  \begin{center}
  \includegraphics[scale=.84]{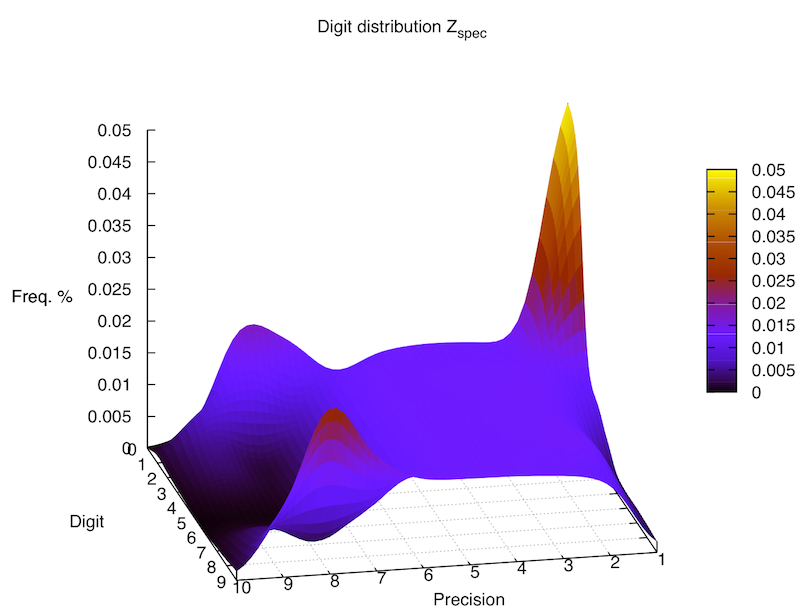}
  \includegraphics[scale=.84]{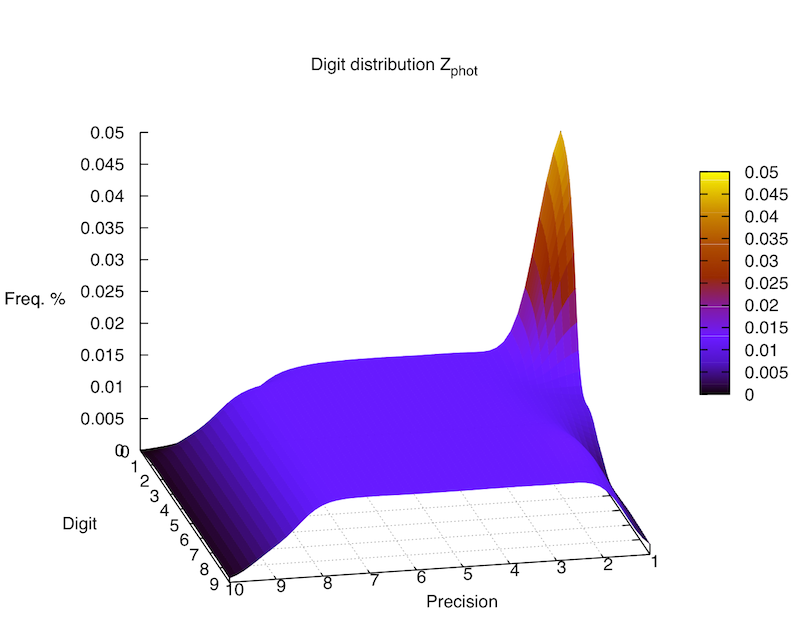}
    \caption{Digit distribution for $z_{spec}$ and $z_{phot}$; Top: Spectrometric digit distribution; Bottom: Photometric digit distribution. Note that digit distribution for $z_{spec}$ has three peaks, but $z_{phot}$ has only one. }
  \label{fig:3D}
  \end{center}
\end{figure}

With this tree it is a trivial task to build bins for data distribution.  Figure~\ref{fig:3D} depicts the frequency distribution for a given digit and precision. There are 100 data points that have been convolved with a Gaussian kernel to produce surface planes in order to assemble three-dimensional plots.

This helps to build a cluster-wise mapping of the data. Following the Figure~\ref{fig:3D} top panel we observe that for the first decimal digit most data observations are concentrated in the digits 0, 1, 2, and 3. Then the rest of decimal precision data is uniformly distributed, gradually going towards zero when the level of precision increases. There is the exception of two peaks, for precision equal to 8. This turns out to be useful because when comparing the $z_{spec}$ and $z_{phot}$ digit distribution we do not find the same peaks in $z_{phot}$.  This is very useful because now we can discriminate which observations are more reliable in  $z_{phot}$ through different characteristics of the data associated with the peaks. 

\clearpage 

\section{Concluding Remarks on the Astronomical Case Study and Other Applications}

In the astronomy case clusters generated with the Baire distance can be useful when calibrating redshifts. In general, applying the Baire method to cases where digit precision is important  can be of relevance, specifically to  highlight data ``bins'' and some of their properties.

Note that when two numbers share 3 prefix digits, and base 10 is used, we have a Baire distance of $\db = 10^{-3}$. We may not need to define the actual (ultra)metric values. It may be, in fact, more convenient to work on the hierarchy, with its different levels.

In section~\ref{subsection:clustering-based-baire} we showed how we could derive that $82.8\%$ of values for $z_{spec}$ and $z_{phot}$ have at least two common prefix digits. This is a powerful result in practice when we recall that we can find very efficiently where these $82.8\%$ of the astronomical objects are. 

Using the Baire distance we showed in section~\ref{section:digit-distribution} that $z_{spec}$ and $z_{phot}$ signals can be stored in a tree like structure. This is advantageous when measuring the digit distribution for each signal.  When comparing these distributions, it can easily be seen where the differences arise.

The Baire distance has proved very useful in a number of cases, for instance 
in~\cite{murtagh08} this distance is used in conjunction with random 
projection~\cite{vempala04} as the basis for clustering 
a large dataset of chemical 
compounds achieving results comparable to $k$-means but with better 
performance due to the lower computational complexity of the Baire-based 
clustering method.

Other application areas include text mining and semantic 
preservation~\cite{Pereira00}.  For more details refer 
to~\cite{pedrophd} where a number of examples are discussed.

\section{Conclusions}

The Euclidean distance is appropriate for real-valued data.  In this work 
we have instead focused on an m-adic (m a non-negative integer) number 
representation.  

In this work the distance called the Baire distance is presented.  This distance has been very recently introduced into data analysis.  We show how this distance can be used to generate clusters in a way that is computationally inexpensive when compared with more traditional techniques. As an ultrametric, the 
distance directly induces a hierarchy.
 Hence the Baire distance lends 
itself very well to the new hierarchical clustering method that we have 
introduced here.  

We presented a case study in this article to motivate 
the approach, more particularly to show how it achieved comparable performance
with respect to k-means, and finally to demonstrate how it greatly outperforms 
k-means (and {\em a fortiori} any traditional hierarchical clustering 
algorithm) computationally.

\begin{appendix}
	\begin{sidewaystable}
\centering
\section{Appendix}\label{appendixA}
\scalebox{0.55}{
\begin{tabular} {|c|cccccccccccccccccccccccccccccccccc|} 
	\hline 	---   &  21 &   1 &   6 &  38 &  25 &  58 &  32 &  20 &  15 &  13 &  14 &  37 &   4 &  17 &   2 &  51 &  30 &  16 &  28 &  44 &  59 &  46 &  23 &  48 &  33 &  60 &  40 &  35 &  50 &  42 &  26 &  31 &  27 & 56
	 \\\hline
	  \hline  015 &3733 &   0 &   0 &   0 &   0 &   0 &   0 &   0 &   0 &   0 &   0 &   0 &   0 &   0 &   0 &   0 &   0 &   0 &   0 &   0 &   0 &   0 &   0 &   0 &   0 &   0 &   0 &   0 &   0 &   0 &   0 &   0 &   0 &  0 	 
	 \\\hline 004 &   0 &3495 &   0 &   0 &   0 &   0 &   0 &   0 &   0 &   0 &   0 &   0 &   0 &   0 &   0 &   0 &   0 &   0 &   0 &   0 &   0 &   0 &   0 &   0 &   0 &   0 &   0 &   0 &   0 &   0 &   0 &   0 &   0 &  0
	 \\\hline 018 &   0 &   0 &2161 &   0 &   0 &   0 &   0 &   0 &   0 &   0 &   0 &   0 &   0 &   0 &   0 &   0 &   0 &   0 &   0 &   0 &   0 &   0 &   0 &   0 &   0 &   0 &   0 &   0 &   0 &   0 &   0 &   0 &   0 &  0
	 \\\hline 020 &   0 &   0 &   0 &1370 &   0 &   0 &   0 &   0 &   0 &   0 &   0 &   0 &   0 &   0 &   0 &   0 &   0 &   0 &   0 &   0 &   0 &   0 &   0 &   0 &   0 &   0 &   0 &   0 &   0 &   0 &   0 &   0 &   0 &  0
	 \\\hline 001 &   0 &   0 &   0 &   0 & 968 &   0 &   0 &   0 &   0 &   0 &   0 &   0 &   0 &   0 &   0 &   0 &   0 &   0 &   0 &   0 &   0 &   0 &   0 &   0 &   0 &   0 &   0 &   0 &   0 &   0 &   0 &   0 &   0 &  0
	 \\\hline 000 &   0 &   0 &   0 &   0 & 515 &   0 &   0 &   0 &   0 &   0 &   0 &   0 &   0 &   0 &   0 &   0 &   0 &   0 &   0 &   0 &   0 &   0 &   0 &   0 &   0 &   0 &   0 &   0 &   0 &   0 &   0 &   0 &   0 &  0
	 \\\hline 022 &   0 &   0 &   0 &   0 &   0 & 896 &   0 &   0 &   0 &   0 &   0 &   0 &   0 &   0 &   0 &   0 &   0 &   0 &   0 &   0 &   0 &   0 &   0 &   0 &   0 &   0 &   0 &   0 &   0 &   0 &   0 &   0 &   0 &  0
	 \\\hline 034 &   0 &   0 &   0 &   0 &   0 &   0 & 764 &   0 &   0 &   0 &   0 &   0 &   0 &   0 &   0 &   0 &   0 &   0 &   0 &   0 &   0 &   0 &   0 &   0 &   0 &   0 &   0 &   0 &   0 &   0 &   0 &   0 &   0 &  0
	 \\\hline 036 &   0 &   0 &   0 &   0 &   0 &   0 &   0 & 652 &   0 &   0 &   0 &   0 &   0 &   0 &   0 &   0 &   0 &   0 &   0 &   0 &   0 &   0 &   0 &   0 &   0 &   0 &   0 &   0 &   0 &   0 &   0 &   0 &   0 &  0
	 \\\hline 037 &   0 &   0 &   0 &   0 &   0 &   0 &   0 & 508 &   0 &   0 &   0 &   0 &   0 &   0 &   0 &   0 &   0 &   0 &   0 &   0 &   0 &   0 &   0 &   0 &   0 &   0 &   0 &   0 &   0 &   0 &   0 &   0 &   0 &  0
	 \\\hline 026 &   0 &   0 &   0 &   0 &   0 &   0 &   0 &   0 & 555 &   0 &   0 &   0 &   0 &   0 &   0 &   0 &   0 &   0 &   0 &   0 &   0 &   0 &   0 &   0 &   0 &   0 &   0 &   0 &   0 &   0 &   0 &   0 &   0 &  0
	 \\\hline 027 &   0 &   0 &   0 &   0 &   0 &   0 &   0 &   0 & 464 &   0 &   0 &   0 &   0 &   0 &   0 &   0 &   0 &   0 &   0 &   0 &   0 &   0 &   0 &   0 &   0 &   0 &   0 &   0 &   0 &   0 &   0 &   0 &   0 &  0
	 \\\hline 032 &   0 &   0 &   0 &   0 &   0 &   0 &   0 &   0 &   0 & 484 &   0 &   0 &   0 &   0 &   0 &   0 &   0 &   0 &   0 &   0 &   0 &   0 &   0 &   0 &   0 &   0 &   0 &   0 &   0 &   0 &   0 &   0 &   0 &  0
	 \\\hline 030 &   0 &   0 &   0 &   0 &   0 &   0 &   0 &   0 &   0 &   0 & 430 &   0 &   0 &   0 &   0 &   0 &   0 &   0 &   0 &   0 &   0 &   0 &   0 &   0 &   0 &   0 &   0 &   0 &   0 &   0 &   0 &   0 &   0 &  0 
	 \\\hline 045 &   0 &   0 &   0 &   0 &   0 &   0 &   0 &   0 &   0 &   0 &   0 & 398 &   0 &   0 &   0 &   0 &   0 &   0 &   0 &   0 &   0 &   0 &   0 &   0 &   0 &   0 &   0 &   0 &   0 &   0 &   0 &   0 &   0 &  0
	 \\\hline 044 &   0 &   0 &   0 &   0 &   0 &   0 &   0 &   0 &   0 &   0 &   0 & 295 &   0 &   0 &   0 &   0 &   0 &   0 &   0 &   0 &   0 &   0 &   0 &   0 &   0 &   0 &   0 &   0 &   0 &   0 &   0 &   0 &   0 &  0
	 \\\hline 047 &   0 &   0 &   0 &   0 &   0 &   0 &   0 &   0 &   0 &   0 &   0 &   0 & 350 &   0 &   0 &   0 &   0 &   0 &   0 &   0 &   0 &   0 &   0 &   0 &   0 &   0 &   0 &   0 &   0 &   0 &   0 &   0 &   0 &  0
	 \\\hline 048 &   0 &   0 &   0 &   0 &   0 &   0 &   0 &   0 &   0 &   0 &   0 &   0 &  57 &   0 &   0 &   0 &   0 &   0 &   0 &   0 &   0 &   0 &   0 &   0 &   0 &   0 &   0 &   0 &   0 &   0 &   0 &   0 &   0 &  0
	 \\\hline 049 &   0 &   0 &   0 &   0 &   0 &   0 &   0 &   0 &   0 &   0 &   0 &   0 &   5 &   0 &   0 &   0 &   0 &   0 &   0 &   0 &   0 &   0 &   0 &   0 &   0 &   0 &   0 &   0 &   0 &   0 &   0 &   0 &   0 &  0
	 \\\hline 050 &   0 &   0 &   0 &   0 &   0 &   0 &   0 &   0 &   0 &   0 &   0 &   0 &   1 &   0 &   0 &   0 &   0 &   0 &   0 &   0 &   0 &   0 &   0 &   0 &   0 &   0 &   0 &   0 &   0 &   0 &   0 &   0 &   0 &  0
	 \\\hline 039 &   0 &   0 &   0 &   0 &   0 &   0 &   0 &   0 &   0 &   0 &   0 &   0 &   0 & 278 &   0 &   0 &   0 &   0 &   0 &   0 &   0 &   0 &   0 &   0 &   0 &   0 &   0 &   0 &   0 &   0 &   0 &   0 &   0 &  0
 	 \\\hline 024 &   0 &   0 &   0 &   0 &   0 &   0 &   0 &   0 &   0 &   0 &   0 &   0 &   0 &   0 & 260 &   0 &   0 &   0 &   0 &   0 &   0 &   0 &   0 &   0 &   0 &   0 &   0 &   0 &   0 &   0 &   0 &   0 &   0 &  0
	 \\\hline 041 &   0 &   0 &   0 &   0 &   0 &   0 &   0 &   0 &   0 &   0 &   0 &   0 &   0 &   0 &   0 & 231 &   0 &   0 &   0 &   0 &   0 &   0 &   0 &   0 &   0 &   0 &   0 &   0 &   0 &   0 &   0 &   0 &   0 &  0
	 \\\hline 042 &   0 &   0 &   0 &   0 &   0 &   0 &   0 &   0 &   0 &   0 &   0 &   0 &   0 &   0 &   0 & 225 &   0 &   0 &   0 &   0 &   0 &   0 &   0 &   0 &   0 &   0 &   0 &   0 &   0 &   0 &   0 &   0 &   0 &  0
	 \\\hline 002 &   0 &   0 &   0 & 247 &   0 &   0 &   0 &   0 &   0 &   0 &   0 &   0 &   0 &   0 &   0 &   0 &1870 &   0 &   0 &   0 &   0 &   0 &   0 &   0 &   0 &   0 &   0 &   0 &   0 &   0 &   0 &   0 &   0 &  0
	 \\\hline 003 &   0 & 523 &   0 &   0 &   0 &   0 &   0 &   0 &   0 &   0 &   0 &   0 &   0 &   0 &   0 &   0 &2320 &   0 &   0 &   0 &   0 &   0 &   0 &   0 &   0 &   0 &   0 &   0 &   0 &   0 &   0 &   0 &   0 &  0
	 \\\hline 005 &   0 & 118 &   0 &   0 &   0 &   0 &   0 &   0 &   0 &   0 &   0 &   0 &   0 &   0 &   0 &   0 &   0 &1720 &2392 &   0 &   0 &   0 &   0 &   0 &   0 &   0 &   0 &   0 &   0 &   0 &   0 &   0 &   0 &  0
	 \\\hline 006 &   0 &   0 &   0 &   0 &   0 &   0 &   0 &   0 &   0 &   0 &   0 &   0 &   0 &   0 &   0 &   0 &   0 &   0 & 389 &6024 &   0 &   0 &   0 &   0 &   0 &   0 &   0 &   0 &   0 &   0 &   0 &   0 &   0 &  0
	 \\\hline 007 &   0 &   0 &   0 &   0 &   0 &   0 &   0 &   0 &   0 &   0 &   0 &   0 &   0 &   0 &   0 &   0 &   0 &   0 &   0 &1417 & 825 &5989 &   0 &   0 &   0 &   0 &   0 &   0 &   0 &   0 &   0 &   0 &   0 &  0
	 \\\hline 008 &   0 &   0 &   0 &   0 &   0 &   0 &   0 &   0 &   0 &   0 &   0 &   0 &   0 &   0 &   0 &   0 &   0 &   0 &   0 &   0 &   0 & 559 &7001 &   0 &   0 &   0 &   0 &   0 &   0 &   0 &   0 &   0 &   0 &  0
	 \\\hline 009 &   0 &   0 &   0 &   0 &   0 &   0 &   0 &   0 &   0 &   0 &   0 &   0 &   0 &   0 &   0 &   0 &   0 &   0 &   0 &   0 &   0 &   0 &7904 &  59 &5042 &   0 &   0 &   0 &   0 &   0 &   0 &   0 &   0 &  0
	 \\\hline 010 &   0 &   0 &   0 &   0 &   0 &   0 &1613 &   0 &   0 &   0 &   0 &   0 &   0 &   0 &   0 &   0 &   0 &   0 &   0 &   0 &   0 &   0 &   0 &   0 &   0 & 710 &3148 &   0 &   0 &   0 &   0 &   0 &   0 &  0
	 \\\hline 011 &   0 &   0 &   0 &   0 &   0 &   0 &   0 &   0 &   0 &   0 &   0 &   0 &   0 &   0 &   0 &   0 &   0 &   0 &   0 &   0 &   0 &   0 &   0 &   0 &   0 &   0 & 213 &5437 &   0 &   0 &   0 &   0 &   0 &  0
	 \\\hline 012 &   0 &   0 &   0 &   0 &   0 &   0 &   0 &   0 &   0 &   0 &   0 &   0 &   0 &   0 &   0 &   0 &   0 &   0 &   0 &   0 &   0 &   0 &   0 &   0 &   0 &   0 &   0 &1784 & 239 &3244 &   0 &   0 &   0 &  0
	 \\\hline 013 &   0 &   0 &   0 &   0 &   0 &   0 &   0 &   0 &   0 &   0 &   0 &   0 &   0 &   0 &   0 &   0 &   0 &   0 &   0 &   0 &   0 &   0 &   0 &   0 &   0 &   0 &   0 &   0 &   0 &4392 & 861 &   0 &   0 &  0
	 \\\hline 014 & 517 &   0 &   0 &   0 &   0 &   0 &   0 &   0 &   0 &   0 &   3 &   0 &   0 &   0 &   0 &   0 &   0 &   0 &   0 &   0 &   0 &   0 &   0 &   0 &   0 &   0 &   0 &   0 &   0 &   0 &3099 & 927 &   0 &  0
	 \\\hline 016 & 697 &   0 &   0 &   0 &   0 &   0 &   0 &   0 &   0 &   0 &   0 &   0 &   0 &   0 &   0 &   0 &   0 &   0 &   0 &   0 &   0 &   0 &   0 &   0 &   0 &   0 &   0 &   0 &   0 &   0 &   0 &   0 & 259 &2153
	 \\\hline 017 &   0 &   0 & 116 &   0 &   0 &   0 &   0 &   0 &   0 &   0 &   0 &   0 &   0 &   0 &   0 &   0 &   0 &   0 &   0 &   0 &   0 &   0 &   0 &   0 &   0 &   0 &   0 &   0 &   0 &   0 &   0 &   0 &   0 &2302
	 \\\hline 019 &   0 &   0 & 644 &1187 &   0 &   0 &   0 &   0 &   0 &   0 &   0 &   0 &   0 &   0 &   0 &   0 &   0 &   0 &   0 &   0 &   0 &   0 &   0 &   0 &   0 &   0 &   0 &   0 &   0 &   0 &   0 &   0 &   0 &  0
	 \\\hline 021 &   0 &   0 &   0 & 245 &   0 & 787 &   0 &   0 &   0 &   0 &   0 &   0 &   0 &   0 &   0 &   0 &   0 &   0 &   0 &   0 &   0 &   0 &   0 &   0 &   0 &   0 &   0 &   0 &   0 &   0 &   0 &   0 &   0 &  0
	 \\\hline 023 &   0 &   0 &   0 &   0 &   0 & 294 & 287 &   0 &   0 &   0 &   0 &   0 &   0 &   0 &   0 &   0 &   0 &   0 &   0 &   0 &   0 &   0 &   0 &   0 &   0 &   0 &   0 &   0 &   0 &   0 &   0 &   0 &   0 &  0
	 \\\hline 035 &   0 &   0 &   0 &   0 &   0 &   0 & 603 & 239 &   0 &   0 &   0 &   0 &   0 &   0 &   0 &   0 &   0 &   0 &   0 &   0 &   0 &   0 &   0 &   0 &   0 &   0 &   0 &   0 &   0 &   0 &   0 &   0 &   0 &  0
	 \\\hline 033 &   0 &   0 &   0 &   0 &   0 &   0 & 129 &   0 &   0 & 463 &   0 &   0 &   0 &   0 &   0 &   0 &   0 &   0 &   0 &   0 &   0 &   0 &   0 &   0 &   0 &   0 &   0 &   0 &   0 &   0 &   0 &   0 &   0 &  0
	 \\\hline 038 &   0 &   0 &   0 &   0 &   0 &   0 &   0 &  72 &   0 &   0 &   0 &   0 &   0 & 317 &   0 &   0 &   0 &   0 &   0 &   0 &   0 &   0 &   0 &   0 &   0 &   0 &   0 &   0 &   0 &   0 &   0 &   0 &   0 &  0
	 \\\hline 025 &   0 &   0 &   0 &   0 &   0 &   0 &   0 &   0 &  20 &   0 &   0 &   0 &   0 &   0 & 334 &   0 &   0 &   0 &   0 &   0 &   0 &   0 &   0 &   0 &   0 &   0 &   0 &   0 &   0 &   0 &   0 &   0 &   0 &  0
	 \\\hline 028 &   0 &   0 &   0 &   0 &   0 &   0 &   0 &   0 & 204 &   0 & 275 &   0 &   0 &   0 &   0 &   0 &   0 &   0 &   0 &   0 &   0 &   0 &   0 &   0 &   0 &   0 &   0 &   0 &   0 &   0 &   0 &   0 &   0 &  0
	 \\\hline 031 &   0 &   0 &   0 &   0 &   0 &   0 &   0 &   0 &   0 & 462 &   3 &   0 &   0 &   0 &   0 &   0 &   0 &   0 &   0 &   0 &   0 &   0 &   0 &   0 &   0 &   0 &   0 &   0 &   0 &   0 &   0 &   0 &   0 &  0
	 \\\hline 029 &   0 &   0 &   0 &   0 &   0 &   0 &   0 &   0 &   0 &   0 & 470 &   0 &   0 &   0 &   0 &   0 &   0 &   0 &   0 &   0 &   0 &   0 &   0 &   0 &   0 &   0 &   0 &   0 &   0 &   0 &   0 &   0 &   0 &  0
	 \\\hline 043 &   0 &   0 &   0 &   0 &   0 &   0 &   0 &   0 &   0 &   0 &   0 & 161 &   0 &   0 &   0 &  76 &   0 &   0 &   0 &   0 &   0 &   0 &   0 &   0 &   0 &   0 &   0 &   0 &   0 &   0 &   0 &   0 &   0 &  0
	 \\\hline 046 &   0 &   0 &   0 &   0 &   0 &   0 &   0 &   0 &   0 &   0 &   0 & 150 & 213 &   0 &   0 &   0 &   0 &   0 &   0 &   0 &   0 &   0 &   0 &   0 &   0 &   0 &   0 &   0 &   0 &   0 &   0 &   0 &   0 &  0
	 \\\hline 040 &   0 &   0 &   0 &   0 &   0 &   0 &   0 &   0 &   0 &   0 &   0 &   0 &   0 &  88 &   0 & 127 &   0 &   0 &   0 &   0 &   0 &   0 &   0 &   0 &   0 &   0 &   0 &   0 &   0 &   0 &   0 &   0 &   0 &  0
	\\\hline
\end{tabular}}
\caption{Cluster comparison based on $\db = 10^{-3}$. Column: $k$-means clusters; Rows: Baire clusters. The array has been row and column permuted in order to highlight the good correspondence.} 
\label{tab:kmeans-60}
\end{sidewaystable}

\end{appendix}
\clearpage
\bibliographystyle{plain}
\bibliography{biblio}

\begin{thebibliography}{10}

\bibitem{Adelman-McCarthy07}
J.~K. Adelman-McCarthy et~al.
\newblock The fifth data release of the {Sloan Digital Sky Survey}.
\newblock {\em The Astrophysical Journal Supplement Series}, 172(2):634--644,
  2007.

\bibitem{benz}
J.-P. Benz\'ecri.
\newblock {\em La Taxinomie}.
\newblock Dunod, Paris, 2nd edition, 1979.

\bibitem{bradley0}
P.~E. Bradley.
\newblock Degenerating families of dendrograms.
\newblock {\em Journal of Classification}, 25:27--42, 2008.

\bibitem{bradley}
P.~E. Bradley.
\newblock Mumford dendrograms.
\newblock {\em Computer Journal}, 53:393--404, 2010.

\bibitem{pedrophd}
P.~Contreras.
\newblock {\em Search and Retrieval in Massive Data Collections}.
\newblock PhD thesis, Royal Holloway, University of London, 2010.

\bibitem{Contreras07}
P.~Contreras and F.~Murtagh.
\newblock Evaluation of hierarchies based on the longest common prefix, or
  {B}aire, metric, 2007.
\newblock {Classification Society of North America (CSNA) meeting, University
  of Illinois. Urbana-Champaign. IL, USA}.

\bibitem{Dabrusco07-1}
R.~D'Abrusco, G.~Longo, M.~Paolillo, M.~Brescia, E.~De~Filippi, A.~Staiano, and
  R.~Tagliaferri.
\newblock The use of neural networks to probe the structure of the nearby
  universe, April 2007.
\newblock \url{http://arxiv.org/pdf/astro-ph/0701137}.

\bibitem{Dabrusco07-2}
R.~D'Abrusco, A.~Staiano, G.~Longo, M.~Brescia, M.~Paolillo, E.~De~Filippis,
  and R.~Tagliaferri.
\newblock {Mining the SDSS archive. I. Photometric redshifts in the nearby
  universe}.
\newblock {\em Astrophysical Journal}, 663(2):752--764, July 2007.

\bibitem{Dabrusco06-1}
R.~D'Abrusco, A.~Staiano, G.~Longo, M.~Paolillo, and E.~De~Filippis.
\newblock {Steps toward a classifier for the virtual observatory. I.
  Classifying the SDSS photometric archive}.
\newblock 1st Workshop of Astronomy and Astrophysics for Students- Naples,
  April 2006.
\newblock \url{http://arxiv.org/abs/0706.4424}.

\bibitem{davey}
B.A. Davey and H.A. Priestley.
\newblock {\em Introduction to Lattices and Order}.
\newblock Cambridge University Press, 2nd edition, 2002.

\bibitem{Fernandez01}
A.~Fern\'andez-Soto, K.~M. Lanzetta, Hsiao-Wen Chen, S.~M. Pascarelle, and
  Noriaki Yahata.
\newblock On the compared accuracy and reliability of spectroscopic and
  photometric redshift measurements.
\newblock {\em The Astrophysical Journal Supplement Series}, 135:41--61, 2001.

\bibitem{ganter}
B.~Ganter and R.~Wille.
\newblock {\em Formal Concept Analysis: Mathematical Foundations}.
\newblock Springer, 1999.
\newblock {{\em Formale Begriffsanalyse. Mathematische Grundlagen}, Springer,
  1996}.

\bibitem{Hartigan79}
J.~A. Hartigan and M.~A. Wong.
\newblock Algorithm {AS} 136: A k-means clustering algorithm.
\newblock {\em Journal of the Royal Statistical Society. Series C (Applied
  Statistics)}, 28(1):100--108, 1979.

\bibitem{seda}
P.~Hitzler and A.~K. Seda.
\newblock The fixed-point theorems of {P}riess-{C}rampe and {R}ibenboim in
  logic programming.
\newblock {\em Fields Institute Communications}, 32:219--235, 2002.

\bibitem{jan0}
M.~F. Janowitz.
\newblock An order theoretic model for cluster analysis.
\newblock {\em SIAM Journal on Applied Mathematics}, 34:55--72, 1978.

\bibitem{jan2010}
M.~F. Janowitz.
\newblock {\em Ordinal and Relational Clustering}.
\newblock World Scientific, 2010.

\bibitem{john}
S.~C. Johnson.
\newblock Hierarchical clustering schemes.
\newblock {\em Psychometrika}, 32:241--254, 1967.

\bibitem{Lerman81}
I.~C. Lerman.
\newblock {\em Classification et Analyse Ordinale des Donn\'ees}.
\newblock Dunod, Paris, 1981.

\bibitem{Longo10}
G.~Longo.
\newblock {DAME}.
\newblock {Data Mining \& Exploration}, 2010.
\newblock \url{http://people.na.infn.it/~astroneural/}.

\bibitem{Murtagh85-1}
F.~Murtagh.
\newblock {\em Multidimensional Clustering Algorithms}.
\newblock Physica-Verlag, 1985.

\bibitem{Murtagh04}
F.~Murtagh.
\newblock On ultrametricity, data coding, and computation.
\newblock {\em {Journal of Classification}}, 21:167--184, September 2004.

\bibitem{Murtagh04-2}
F.~Murtagh.
\newblock Quantifying ultrametricity.
\newblock In J.~Antoch, editor, {\em COMPSTAT 2004 -- Proceedings in
  Computational Statistics}, pages 1561--1568, Prague, Czech Republic, 2004.
  Springer.

\bibitem{Murtagh04-1}
F.~Murtagh.
\newblock Thinking ultrametrically.
\newblock In D.~Banks, L.~House, F.~R. McMorris, P.~Arabie, and W.~Gaul,
  editors, {\em Classification, Clustering, and Data Mining Applications.
  Proceedings of the Meeting of the International Federation of Classification
  Societies (IFCS)}, pages 3--14, Illinois Institute of Technology, Chicago,
  July 2004. Springer.

\bibitem{Murtagh05}
F.~Murtagh.
\newblock Identifying the ultrametricity of time series.
\newblock {\em The European Physical Journal B}, 43(4):573--579, February 2005.

\bibitem{steklov}
F.~Murtagh.
\newblock Symmetry in data mining and analysis: a unifying view based on
  hierarchy.
\newblock {\em Proceedings of Steklov Institute of Mathematics}, 265:177--198,
  2009.

\bibitem{murtagh08}
F.~Murtagh, G.~Downs, and P.~Contreras.
\newblock Hierarchical clustering of massive, high dimensional data sets by
  exploiting ultrametric embedding.
\newblock {\em SIAM Journal on Scientific Computing}, 30(2):707--730, February
  2008.

\bibitem{Pereira00}
J.~Pereira, F.~Schmidt, P.~Contreras, F.~Murtagh, and H.~Astudillo.
\newblock Clustering and semantics preservation in cultural heritage
  information spaces.
\newblock In {\em RIAO'2010, 9th International Conference on Adaptivity,
  Personalization and Fusion of Heterogeneous Information}, pages 100--105,
  Paris, France, 2010.

\bibitem{SDSS}
SDSS.
\newblock {Sloan Digital Sky Survey}, 2008.
\newblock \url{http://www.sdss.org}.

\bibitem{sedacj}
A.~K. Seda and P.~Hitzler.
\newblock Generalized distance functions in the theory of computation.
\newblock {\em Computer Journal}, 53:443--464, 2010.

\bibitem{Rooij78}
A.~C.~M. van Rooij.
\newblock {\em Non-Archimedean Functional Analysis}.
\newblock Marcel Dekker, 1978.

\bibitem{vempala04}
S.~S. Vempala.
\newblock {\em The Random Projection Method. DIMACS: Series in Discrete
  Mathematics and Theoretical Computer Science}, volume~65.
\newblock American Mathematical Society, 2004.

\end{thebibliography}

\end{document}